 \documentclass[pmlr,twocolumn,10pt]{jmlr} % W&CP article
 \jmlrpages{}
\usepackage{array}
\usepackage{lipsum} % For dummy text, remove in actual document

\usepackage{afterpage} % For reverting changes after the first page

\usepackage{booktabs}
\usepackage{siunitx}

\usepackage[switch]{lineno}

\newcommand{\equal}[1]{{\hypersetup{linkcolor=black}\thanks{#1}}}

\theorembodyfont{\upshape}
\theoremheaderfont{\scshape}
\theorempostheader{:}
\theoremsep{\newline}

\jmlrworkshop{Machine Learning for Health (ML4H) 2024} % W&CP title

\title[Evaluating GPT for Identifying Cognitive Impairment in EHR]{Evaluating GPT's Capability in Identifying Stages of Cognitive Impairment from Electronic Health Data}

\author{%
\Name{Yu Leng}\equal{These authors contributed equally} \Email{yleng2@mgh.harvard.edu}\\
\Name{Yingnan He}\footnotemark[1] \Email{yihe1@mgh.harvard.edu}\\
\Name{Colin Magdamo}\Email{cmagdamo@mgh.harvard.edu}\\
\Name{Ana-Maria Vranceanu}\Email{vranceanu@mgh.harvard.edu}\\
\Name{Christine S. Ritchie}\Email{csritchie@mgh.harvard.edu}\\
\Name{Shibani S. Mukerji}\Email{smukerji@mgb.org}\\
\Name{Lidia M. V. R. Moura}\Email{lidia.moura@mgh.harvard.edu}\\
\Name{John R. Dickson}\Email{john.dickson@mgh.harvard.edu}\\
\Name{Deborah Blacker}\Email{dblacker@mgh.harvard.edu}\\
\Name{Sudeshna Das} \Email{sdas5@mgh.harvard.edu}
}

\begin{document}

\maketitle

\begin{abstract}
Identifying cognitive impairment within electronic health records (EHRs) is crucial not only for timely diagnoses but also for facilitating research. Information about cognitive impairment often exists within unstructured clinician notes in EHRs, but manual chart reviews are both time-consuming and error-prone. To address this issue, our study evaluates an automated approach using zero-shot GPT-4o to determine stage of cognitive impairment in two different tasks. First, we evaluated the ability of GPT-4o to determine the global Clinical Dementia Rating (CDR) on specialist notes from 769 patients who visited the memory clinic at Massachusetts General Hospital (MGH), and achieved a weighted kappa score of 0.83. Second, we assessed GPT-4o's ability to differentiate between normal cognition, mild cognitive impairment (MCI), and dementia on all notes in a 3-year window from 860 Medicare patients. GPT-4o attained a weighted kappa score of 0.91 in comparison to specialist chart reviews and 0.96 on cases that the clinical adjudicators rated with high confidence. Our findings demonstrate GPT-4o's potential as a scalable chart review tool for creating research datasets and assisting diagnosis in clinical settings in the future.
\end{abstract}
\begin{keywords}
GPT, EHR, MCI, CDR
\end{keywords}

\paragraph*{Data and Code Availability}
We will share our GitHub repository for the camera-ready version of the paper. Due to the presence of PHI, the data cannot be shared publicly. 

\paragraph*{Institutional Review Board (IRB)}
Relevant ethics approval information will be provided if the paper is accepted.

\section{Introduction}
\label{sec:intro}
Alzheimer’s Disease and Related Dementias (ADRD, referred to hereafter as dementia) describes a group of related neurodegenerative disorders that affect over 6 million people over age 65 in the United States and represent a large and growing problem in the 21st century \citep{RN1}. Timely diagnosis of dementia is crucial for interventions and treatment plans that can help manage symptoms and improve the quality of life for persons living with dementia and their families \citep{RN3}.  Yet, dementia remains under-recognized, under-diagnosed, and under-reported in healthcare records \citep{RN4}. Automated mining of clinical notes have the potential to facilitate clinical diagnosis as well as research studies of dementia.

The Electronic Health Record (EHR)—which includes detailed health history, clinical notes, and other health-system interaction information—offers readily available data and great potential for identifying cognitive impairment in patients without a formal diagnosis in EHR. Despite the prevalence of cognitive impairment data within EHR, these critical insights are often buried in unstructured clinician notes and not readily accessible for clinical decision-making or research. Traditional methods for extracting this information involve labor-intensive manual reviews, which are not only time-consuming but also prone to inconsistencies and errors. To address this gap, several prior studies have applied natural language processing (NLP) and/or large-language models (LLMs) to detect cognitive impairment in clinical notes within EHR, for example \citet{RN11}

However, to our knowledge none of the prior efforts have applied the latest GPT models to this problem. This study introduces and evaluates the use GPT-4o to automate the extraction and interpretation of cognition data from EHRs. We evaluated GPT-4o in two different studies. First, we use GPT-4o to assign a global CDR score on specialist notes (with detailed cognitive evaluation) on patients who visited the memory clinic at Massachusetts General Hospital (MGH). These memory specialist notes have detailed information on cognitive evaluation; our goal was to evaluate whether we could automatically create structured datasets of patient global CDR scores. Second, we evaluated GPT-4o’s performance in assessing stage of cognitive impairment (normal cognition (NC), mild cognitive impairment (MCI), and dementia) in a Medicare patient group using all notes spanning a 3-year period. Here, the motivation was to compare automated to manual chart reviews for either research or clinical diagnosis.

\section{Datasets and Processing}

For the first study, the dataset comprised of 769 latest visit notes of 769 unique patients from the memory clinic at MGH from February 2016 to July 2019. (\tableref{tab:memdemo}). These patients consented to be part of a registry which recorded the global CDR score and diagnoses at their visit, along with other data. Any sentence with mention of CDR  was redacted from the notes using regex before evaluation by GPT-4o. GPT-4o was prompted to assign a global CDR score. 

For the second study, the dataset consisted of a sample of 860 Medicare fee-for-service patients from a previous study by \citet{RN12}.  Each patient’s EHR data between 01/01/2016 – 12/31/2018 was reviewed by an expert physician to label patients with the stage of cognitive impairment (Normal, Normal-to-MCI, MCI, MCI-to-dementia, and dementia) and to assign a confidence level of 1 (lowest) to 4 (highest). The MCI-to-dementia patients were included in the MCI category and Normal-to-MCI were excluded to get three final categories: NC, MCI, and dementia. For this dataset (\tableref{tab:mourademo}), we prepared a summary of summaries with GPT-4o. For each patient, we aggregated outpatient visit notes in chronological order. The context of the note (i.e., the date, department, specialty) was added to each note and summarized by GPT-4o. The notes summaries were then combined chronologically into one document. For each patient, GPT-4o was prompted to generate a "summary of summaries" and make a final diagnosis based on this summary of summaries, classifying the patient’s cognitive status as: NC, MCI, or Dementia.

\begin{table*}[htbp]
 % The first argument is the label.
 % The caption goes in the second argument, and the table contents
 % go in the third argument.
\floatconts
  {tab:memdemo}%
  {\caption{MGH Memory Clinic Patient Demographics (Study I)}}%
  {\begin{tabular}{ccccccc}
  \toprule
      & & \multicolumn{5}{c}{Global CDR Score}\\
      \cmidrule(lr){3-7} % Removed this line to eliminate the horizontal rule
      & Total N & 0 & 0.5 & 1 & 2 & 3\\
      Characteristics & N=769 & N=38 (5\%) & N=267 (35\%) & N=218 (28\%) & N=179 (23\%) & N=67 (9\%)\\
        \midrule
      Age (Mean, SD) & 77.9 (8.0) & 72.3 (9.5) & 76.6 (7.5) & 78.3 (7.7) & 80.2 (7.4) & 79.1 (8.6)\\
 Female & 397 (52\%) & 21 (55\%) & 111 (42\%) & 120 (55\%) & 109 (61\%) & 36 (54\%)\\
 Male & 372 (48\%) & 17 (45\%) & 156 (58\%) & 98 (45\%) & 70 (39\%) & 31 (46\%)\\
 \bottomrule
 \end{tabular}}
\end{table*}

\begin{table}[htbp]
\floatconts
  {tab:mourademo}%
  {\caption{Medicare Patient Demographics (Study II)}}%
  {\begin{tabular}{>{\centering\arraybackslash}m{2cm} >{\centering\arraybackslash}m{1.35cm} >{\centering\arraybackslash}m{1.35cm} >{\centering\arraybackslash}m{1.35cm}}
  \toprule
      &\multicolumn{3}{c}{Cognitive Impairment Stage}\\
      \cmidrule(lr){2-4}
        % \midrule
      Total N&Normal&MCI&Dementia\\
      N=860
&N=530
&N=106
&N=224
\\
Characteristics&(62\%)&(12\%)&(26\%)\\
        \midrule
      Age\newline(Mean, SD)&75.8 
(6.5)&78.2 
(6.6)&83.1 
(7.5)\\
 Female& 304
(57\%)& 53
(50\%)& 150
(67\%)\\
 Male& 226
(43\%)& 53
(50\%)& 74
(33\%)\\
 \bottomrule
 \end{tabular}}
\end{table}

\section{Methodology}

In this study, we evaluated the capability of GPT-4o to identify stage of cognitive impairment by implementing a series of prompt engineering techniques and a Retrieval-Augmented Generation (RAG) approach in the two datasets described above.

\subsection{GPT, Prompt Engineering and Retrieval-Augmented Generation}

For Study I, we tried three approaches. The first approach utilized a structured answer template to guide GPT-4o's analysis of patient visit notes. The response format asks GPT-4o to provide observations and summaries across six key domains in the CDR scoring system \citep{RN18}: i) Memory, ii) Orientation, iii) Judgment and Problem Solving, iv) Community Affairs, v) Home and Hobbies, and vi) Personal Care. The model was required to conclude with an explicit CDR score. This structured format helped standardize the output and ensured that each relevant domain was systematically considered before making the final decision. 

Second, to further enhance GPT-4o's ability to determine stages of cognitive impairment, we implemented a Retrieval-Augmented Generation (RAG) approach. We extracted information from the NACC UDS v3 CDR Dementia Staging Instrument \citep{RN16, RN18}, chunked them into manageable pieces, and indexed them for efficient retrieval. When processing patient visit notes, the model searches for the top three chunks that are most similar to the notes. These relevant pieces of information are then used to augment GPT-4o, providing the model with domain-specific guidance from human experts. 

Third, we asked GPT-4o to include a self-assessment of confidence and a count of explicitly mentioned domains. The model was asked to review the visit notes with a focus on identifying information within the six specific CDR domains and summarize the number of domains with explicit information. Based on the clarity and consistency of the evidence across these domains, GPT-4o was asked to assign a confidence level (low, medium, or high). This method aimed to enhance the reliability of the predictions by integrating a self-evaluation component into the model’s decision-making process.

For Study II \citep{RN12}, based on the summary of summaries from the 3-year window, we asked GPT to provide an overall classification of the patient's cognitive status over the three-year period into one of the following syndromic diagnoses: i) NC, ii) MCI, iii) Dementia. We reasoned that a RAG approach is not required for this simpler task. We also asked GPT-4o to provide a rationale for the classification as well as a confidence level on a 1-100 scale. To ensure consistency between the clinical and GPT-4o ratings, we then used quantile mapping to convert these back to the original 1-4 scale.

\subsection{Evaluation}

To assess the accuracy and reliability of decisions made by GPT, we used several analytical methods. We treated the staging task as an ordinal classification problem, and used quadratic weighted Cohen’s kappa score as the evaluation metric. We also created confusion matrices and computed stratified performance metrics based on GPT confidence levels.

\section{Results}
\begin{figure*}[t!]
\floatconts
  {fig:figure1}
  {\caption{GPT-4o Performance on Two Studies (Top Row: Study I; Bottom Row: Study II)}}
  {%
    \subfigure[GPT with Structured Guidance]{\label{fig:figure1a}%
      \includegraphics[width=0.3\linewidth]{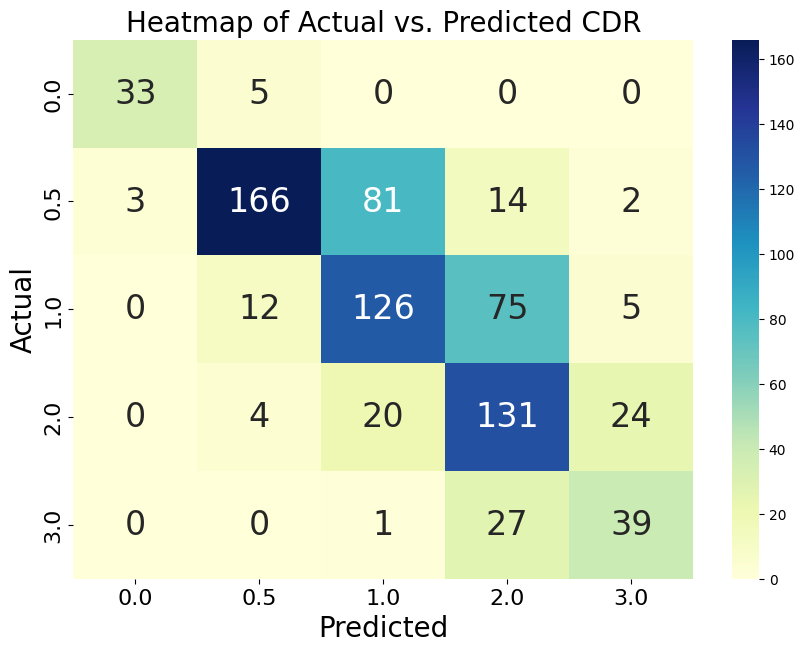}}%
    \qquad
    \subfigure[RAG-Enabled GPT]{\label{fig:figure1b}%
      \includegraphics[width=0.3\linewidth]{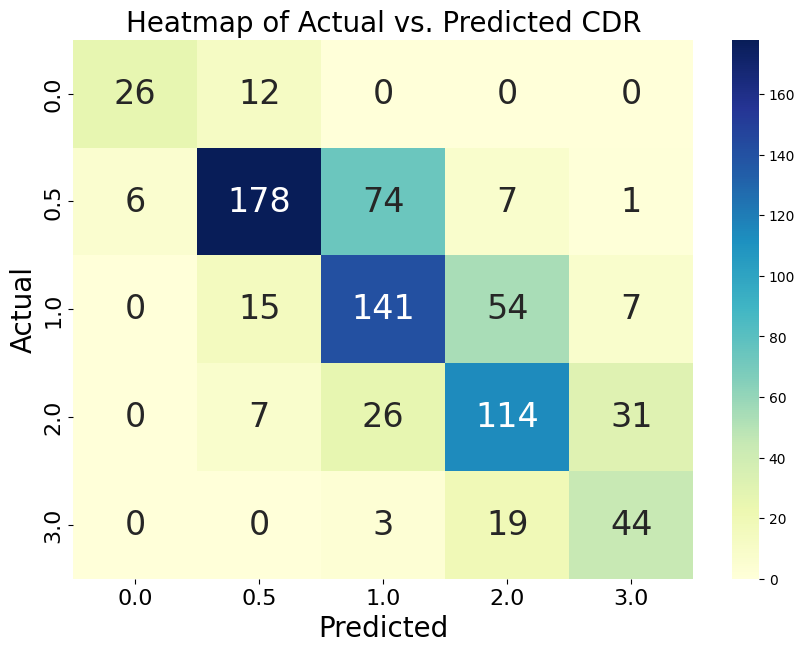}}%
    \qquad
    \subfigure[GPT with Confidence Level and Domain Counts]{\label{fig:figure1c}%
      \includegraphics[width=0.3\linewidth]{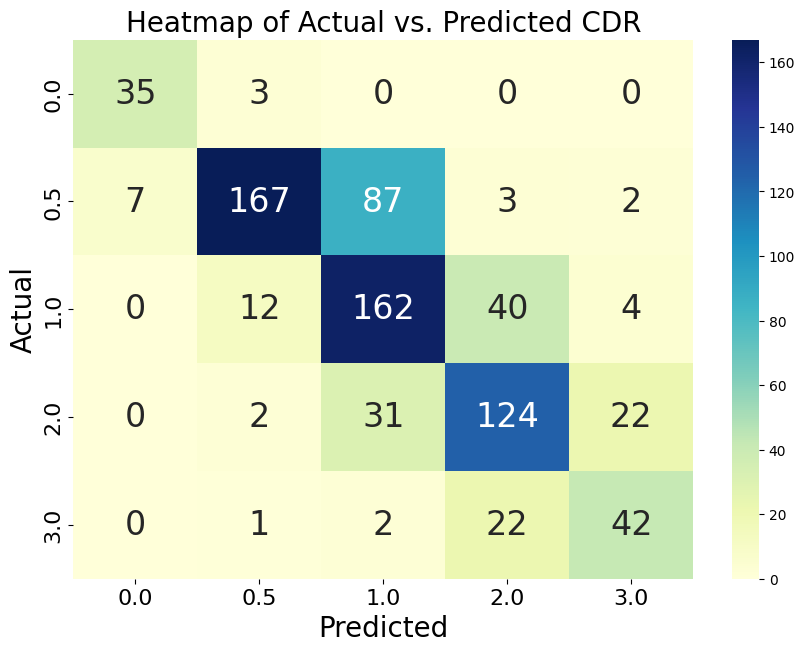}}%
    \qquad
    \subfigure[Heatmap of Actual vs. Predicted Syndromic Dx]{\label{fig:figure2a}%
      \includegraphics[width=0.3\linewidth]{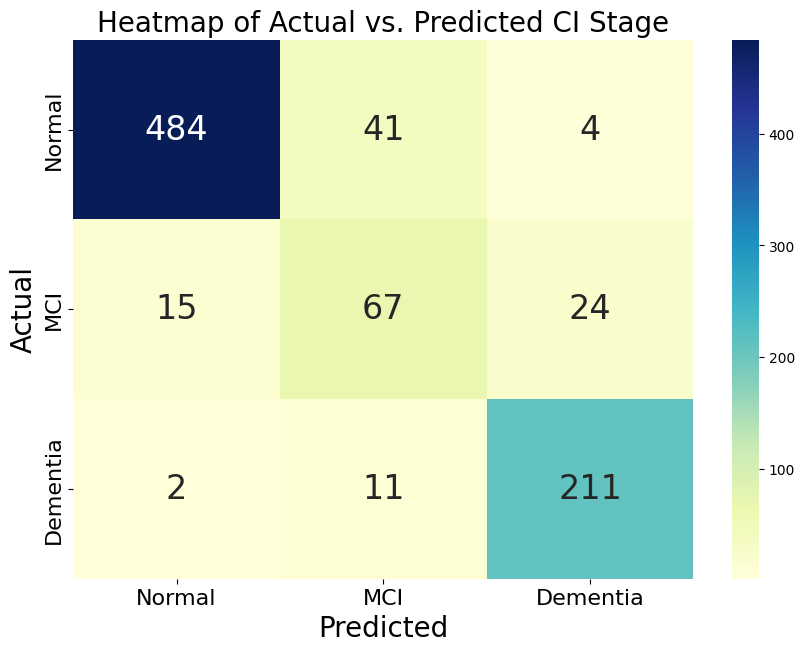}}%
    \qquad
    \subfigure[Weighted Kappa Scores by Physician Confidence Level]{\label{fig:figure2b}%
      \includegraphics[width=0.3\linewidth]{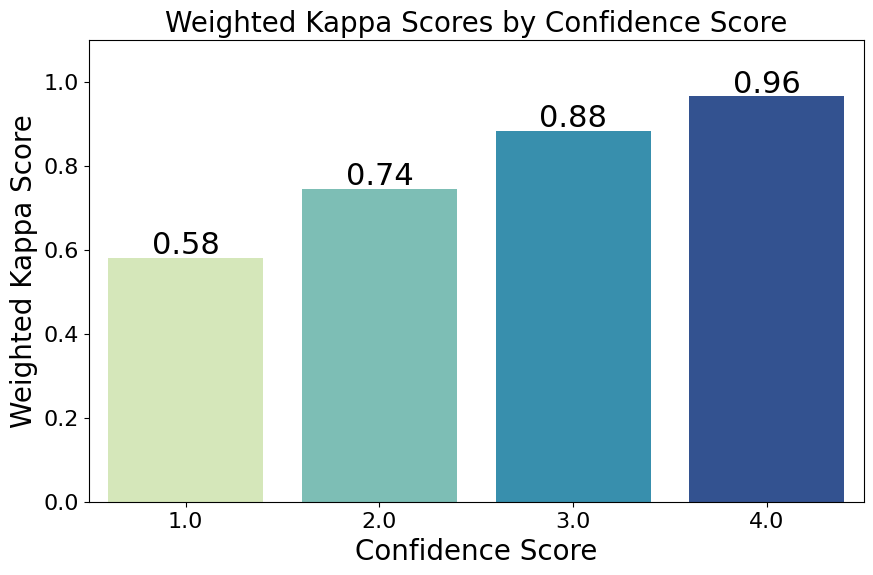}}%
    \qquad
    \subfigure[Confidence Level: Physician vs. GPT-4o]{\label{fig:figure3}%
      \includegraphics[width=0.3\linewidth]{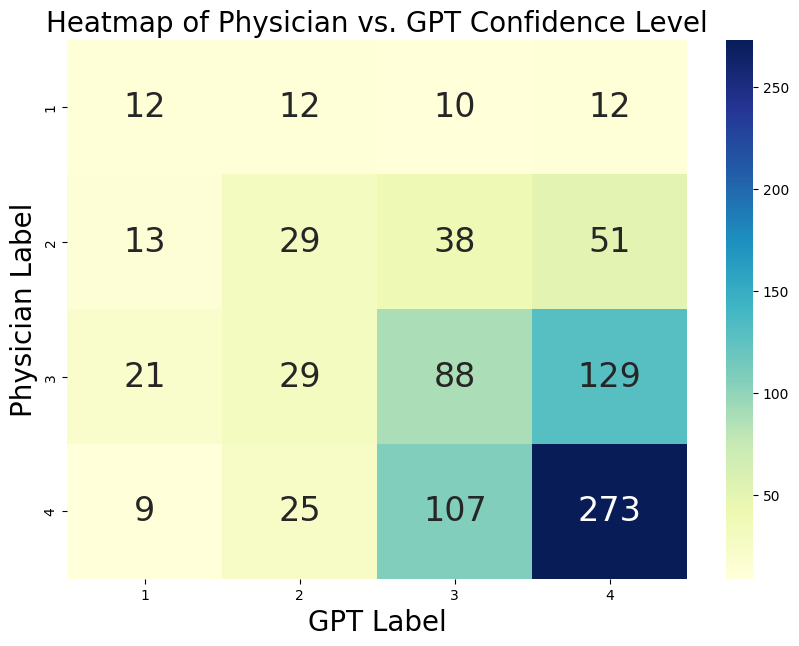}}%
  }
\end{figure*}

For Study I, GPT-4o with structured answer template prompt, RAG-enabled GPT-4o, and GPT-4o with confidence level and domain count prompt achieved weighted Cohen’s kappa scores of 0.79, 0.80, and 0.83 respectively (\figureref{fig:figure1a,fig:figure1b,fig:figure1c}). GPT-4o consistently predicted a higher stage of cognitive impairment than the actual condition in all models. Stratification analysis showed that decisions made with high confidence had the highest weighted Cohen’s kappa, whereas predictions with low or medium confidence had lower values, as expected (low: 0.40; medium: 0.56; high: 0.84). Notably, GPT-4o was ``overconfident", with more than two-thirds of the predictions (618) rated as high confidence, and only a few (5) rated as low. The prompts for the third approach (confidence level and domain count) and a sample output are shown in \appendixref{apd:first}.

For Study II, the overall weighted kappa score is 0.91 (\figureref{fig:figure2a}). A stratification analysis based on the clinical adjudicator’s confidence levels revealed a clear trend: cases adjudicated with higher confidence by physicians demonstrated stronger alignment between GPT-4o's predictions and the physician’s diagnosis (\figureref{fig:figure2b}). This trend indicates that cases rated with higher confidence by physicians were also those where GPT-4o performs exceptionally well. \figureref{fig:figure3} displays the confusion matrix between physician's confidence level in the adjudication and GPT-generated confidence levels. There was strong agreement between physicians and GPT-4o at the highest confidence level, indicating that GPT likely shares a similar understanding with physicians of what the highest confidence level represents. The GPT-4o prompt and sample output is shown in \appendixref{apd:second}. 

\section{Conclusion and Future Work}
GPT-4o demonstrated good performance (weighted kappa 0.79-0.83) on assessing global CDR from the MGH memory clinic visit notes.  Notably, the highest
weighted Cohen’s kappa score was achieved through
the use of prompt engineering techniques that incorporated confidence levels and a count of the documented domains. However, even this approach had several errors (\figureref{fig:figure1c}). The mismatch between the global CDR assigned by the physician and the GPT-determined CDR may stem from some physicians estimating a ``gestalt CDR" based on their overall impression, rather than using the formal scoring algorithm.  The use of RAG-enabled GPT-4o did not significantly enhance the scores in this study; it is possible that GPT-4o possesses sufficient knowledge to determine the stage of cognitive impairment without additional augmentation. In short, our findings on the memory clinic notes indicate that GPT-4o can be used by researchers to create structured datasets—such as those of disease progression—from memory clinic notes, applying manual review to cases rated with low or medium confidence. Such a real-world dataset can serve as a valuable resource for a wide range of dementia studies. 

On the Medicare fee-for-service patients notes, GPT-4o demonstrated even stronger performance (weighted kappa 0.91), perhaps because this task was simpler than scoring a global CDR. The CDR is a detailed measure of cognitive and functional performance across six domains, while staging broadly categorizes cognitive status. Our results underscore GPT's potential for automated chart reviews, and facilitating diagnosis in clinical settings. However, it is important to acknowledge that, as previously reported, there are sociodemographic biases in access to specialists, healthcare utilization, reporting of symptoms, and documentation in clinic notes \citep{RN13,RN14,RN15} that this study does not address. Future work is essential to mitigate these biases in EHR data before they can be deployed at scale.  Additionally, larger studies at multiple healthcare institutions are required to validate GPT as a tool for dementia chart reviews, and to investigate whether GPT-assisted cognitive diagnoses in clinical settings can influence patient outcomes.

\clearpage

\section{Citations and Bibliography}
\bibliography{jmlr-sample}

% Switch to one-column mode for the appendix
\onecolumn
\appendix

\section{GPT-4o prompt and sample answer for memory clinic notes}\label{apd:first}
\begin{figure*}[htbp]
\floatconts
  {fig:appendixa}%
  %{\caption{}}%
  {}
  {\includegraphics[width=0.9\linewidth]{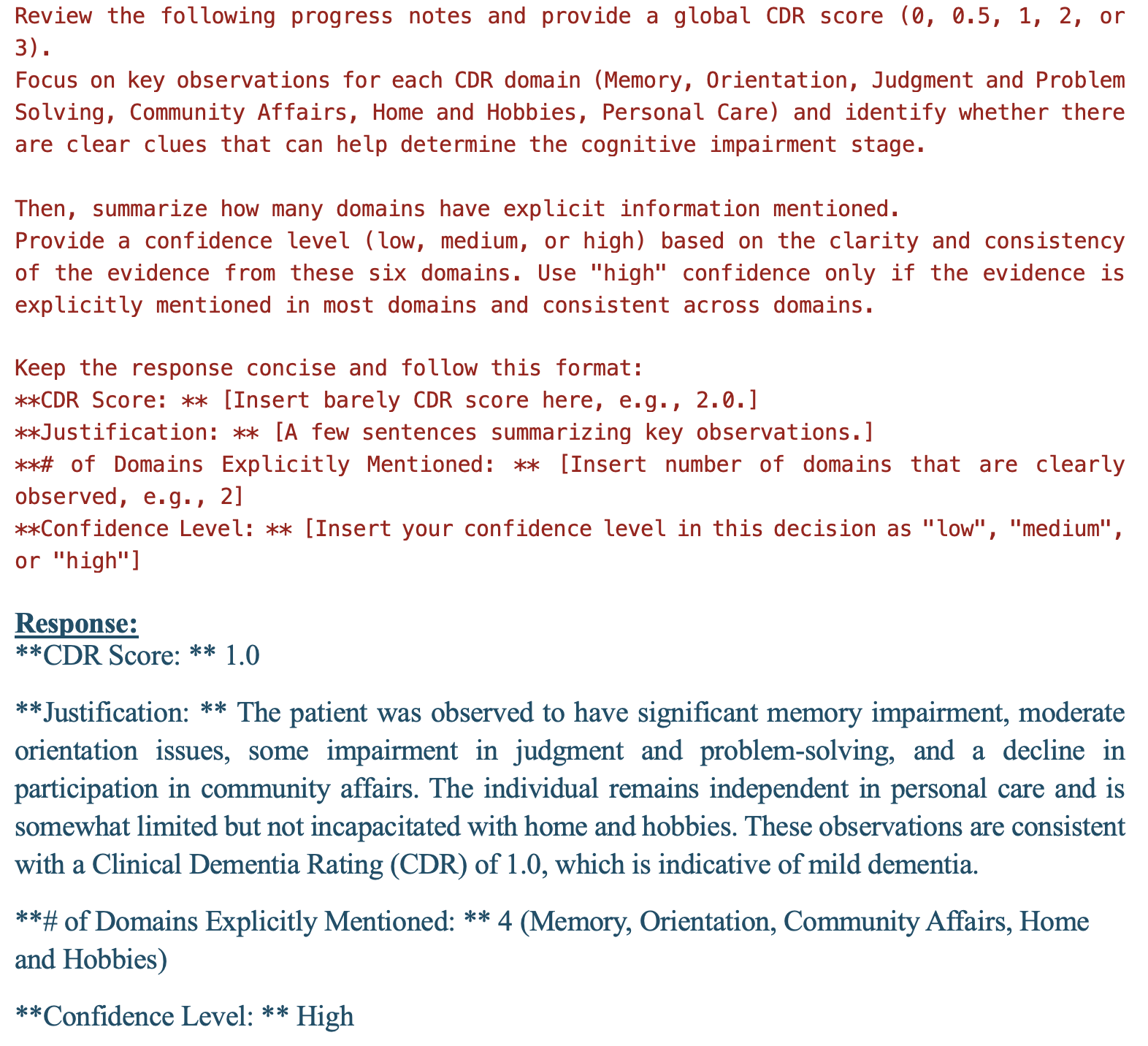}}
\end{figure*}

\newpage
\section{GPT-4o prompt and sample answer for Medicare patient dataset}\label{apd:second}
\begin{figure*}[htbp]
\floatconts
  {fig:appendixb}%
  %{\caption{}}%
  {}
  {\includegraphics[width=0.9\linewidth]{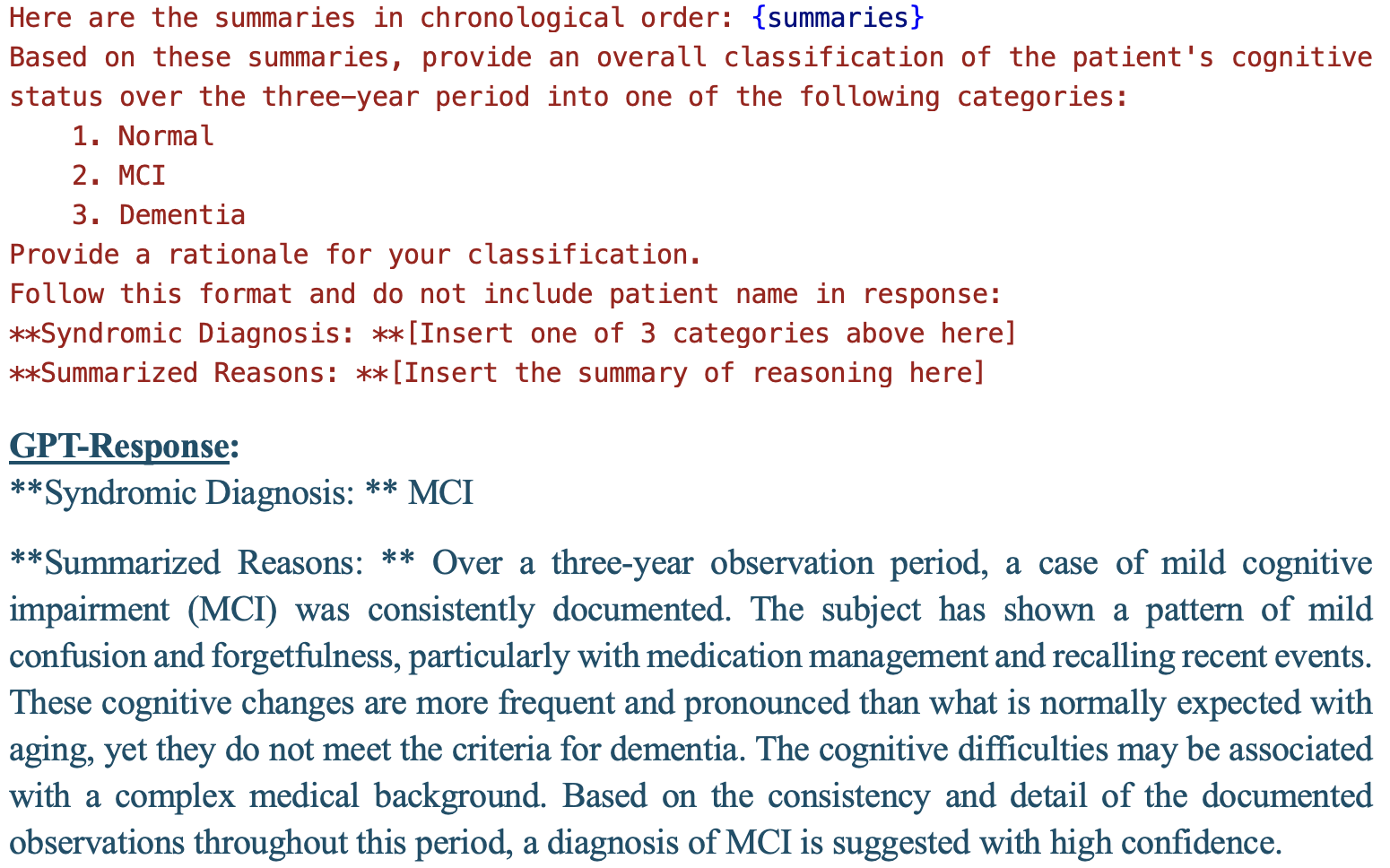}}
\end{figure*}

\end{document}